\documentclass[11pt]{article}

\usepackage[preprint]{acl}

\usepackage{times}
\usepackage{latexsym}

\usepackage[T1]{fontenc}
\usepackage[utf8]{inputenc}

\usepackage{microtype}

\usepackage{inconsolata}

\usepackage{graphicx}

\usepackage{times}
\usepackage{latexsym}
\usepackage{makecell}
\usepackage{enumitem}
\usepackage{amssymb}
\usepackage{textcomp}

\usepackage[T1]{fontenc}
\usepackage[utf8]{inputenc}

\usepackage{microtype}

\usepackage{inconsolata}

\usepackage{graphicx}
\usepackage[table]{xcolor}
\definecolor{ardiff}{RGB}{255, 232, 231}
\definecolor{aronly}{RGB}{217, 229, 240}
\definecolor{diff}{RGB}{220,245,220}

\usepackage{hyperref}
\usepackage{url}
\usepackage{booktabs}       % professional-quality tables
\usepackage{amsfonts}       % blackboard math symbols
\usepackage{nicefrac}       % compact symbols for 1/2, etc.
\usepackage{multirow}
\usepackage{amsmath}
\usepackage{colortbl}
\usepackage{algorithm}
\usepackage{algpseudocode}
\usepackage{arydshln}
\definecolor{LightCyan}{rgb}{0.93,0.95,1}
\newcolumntype{b}{>{\columncolor{LightCyan}}c}
\usepackage{minitoc} % Package for local table of contents

\title{Evo: Autoregressive–Diffusion Large Language Models with Evolving Balance}

\author{Junde Wu \\
  University of Oxford \\\And
  Minhao Hu \\
  University of Oxford \\\And
  Jiayuan Zhu \\
  University of Oxford \\\And
  Yuyuan Liu \\
  University of Oxford \\\AND
  Tianyi Zhang \\
  National University of Singapore \\\And
  Kang Li \\
  University of Oxford \\\And
  Jingkun Chen \\
  University of Oxford \\\AND
  Jiazhen Pan \\
  Technical University of Munich \\\And
  Min Xu \\
 Carnegie Mellon University \\\And
  Yueming Jin \\
  National University of Singapore
  }

\begin{document}
\maketitle
\begin{abstract}
We introduce \textbf{Evo}, a duality latent trajectory model that bridges autoregressive (AR) and diffusion-based language generation within a continuous evolutionary generative framework. Rather than treating AR decoding and diffusion generation as separate paradigms, Evo reconceptualizes text generation as a latent flow: each token is associated with a vector-valued embedding that evolves over a progression variable $t_i \in [0, 1]$, indicating its semantic maturity. Low $t_i$ values correspond to confident AR-like refinement, while high values invoke diffusion-style planning, allowing the model to adaptively balance AR and diffusion based on uncertainty. Theoretically, we show that both AR and diffusion models emerge as discretizations of a shared probability flow, and we derive Evo’s training objective from a unified variational ELBO. The model is implemented as a time-conditioned Transformer governed by a shared vector field, trained end-to-end to jointly infer latent codes and their progression times. During decoding, Evo performs efficient, semantics-aware refinement, achieving high-quality outputs without sacrificing speed. Empirically, Evo 8B achieves state-of-the-art or highly competitive results on 15 diverse benchmarks, including reasoning (GSM8K, ARC-C), code generation (HumanEval, MBPP), and general language understanding, while maintaining fast inference speed. Our results demonstrate that Evo delivers a new paradigm for LLM design with strong generation quality, robust symbolic reasoning, and decoding efficiency. 
\end{abstract}

\addtocontents{toc}{\protect\setcounter{tocdepth}{-10}}

\section{Introduction}
Large Language Models (LLMs) have emerged as a transformative paradigm in natural language processing, demonstrating remarkable capabilities in open-ended generation, reasoning, and multi-turn dialogue\cite{wu2025agentic, wei2022chain}. From GPT-series to instruction-tuned variants like ChatGPT and Claude, the predominant architecture driving these successes is the autoregressive (AR) Transformer, which models the conditional distribution of a sequence token-by-token in a left-to-right manner\cite{radford2018gpt, radford2019gpt2}. This causal decoding framework has enabled scalable pretraining, fine-tuned alignment, and effective sampling. Yet, despite their ubiquity, AR-based LLMs exhibit the limitations that they operate in a strictly unidirectional generation regime and often exhibit compounding errors due to greedy or approximate decoding heuristics.

To address these challenges, a growing body of work has explored diffusion-based LLMs, generative models that synthesize text by iteratively denoising corrupted inputs\cite{zhu2025llada, gong2022diffuseq}. Inspired by their success in image and audio domains, diffusion models for language seek to replace one-pass AR decoding with a multi-step process that allows for non-sequential token generation. These models offer promising benefits, such as iterative self-correction and better global coordination\cite{wu2019universal, xing2023diff, wu2022medsegdiff, amit2021segdiff, kim2022diffusion, wu2022fat}. However, diffusion models often require extensive inference steps, lack explicit control over high-level semantics, and generally underperform AR models in perplexity due to lossy training objectives and the absence of strategic planning.

\begin{figure*}[t]
\centering
\includegraphics[width=0.95\linewidth]{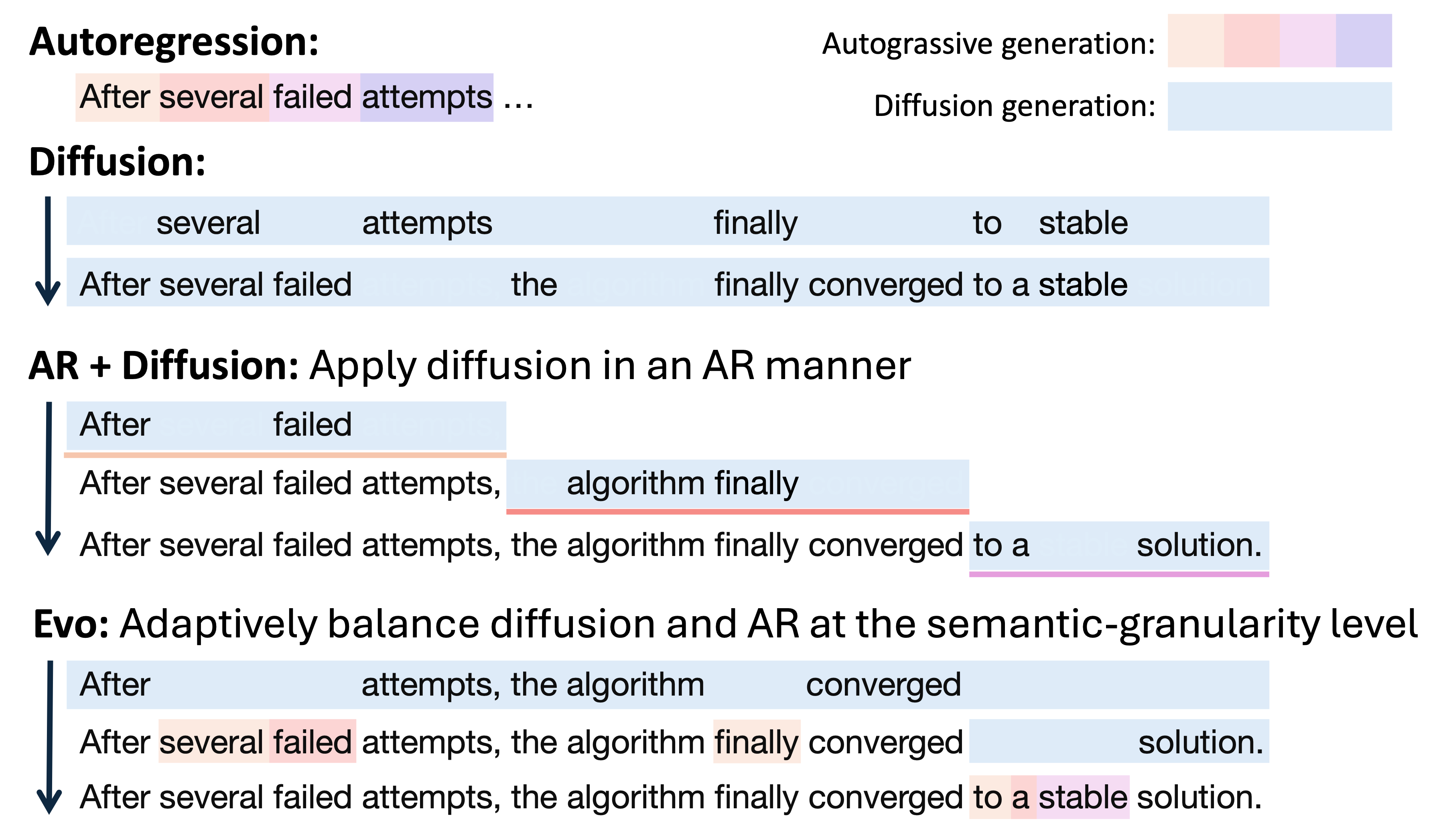}
\vspace{-5pt}
\caption{ Evo generates text by evolving a diffusion-based semantic scaffold into token-level autoregressive realizations along a shared latent trajectory, allowing different tokens to reach semantic maturity at different times. Unlike prior autoregressive, diffusion, and AR+diffusion models, which either generate strictly left-to-right or apply diffusion in a fixed autoregressive order, Evo adaptively balances diffusion-style global planning and autoregressive next-token prediction by semantic granularity, selectively refining mature tokens while continuing to finalize the remaining scaffold within the same sentence.}
\label{fig:cover}
\vspace{-10pt}
\end{figure*}

In this paper, we introduce Evo, a duality latent flow model that generalizes autoregressive (AR) and diffusion-based language generation within a single, continuous generative framework. Different from existing AR–diffusion hybrids, we take autoregressive and diffusion generation as two manifestations of the same underlying diffusion process at different progression times in latent space. This inherent duality directly motivates Evo’s unique design, allowing it to adaptively modulate the balance between AR-style refinement and diffusion-style planning based on the desired level of generation granularity, thereby jointly improving semantic clarity and detail richness. An illustration is shown in Fig. \ref{fig:cover}.

Specifically, at the core of our framework lies a theoretical unification of AR and diffusion. We show both are shown to emerge as discretizations of a shared probabilistic flow. We formalize this connection by defining latent trajectories governed by a time-indexed vector field, and demonstrate that AR generation corresponds to deterministic flows near the origin, while diffusion-based decoding arises as stochastic score-following in the limit. This continuous perspective allows Evo to model generation as a structured path over latent states, with token-level progression times ${t_i}$ that control how far each representation lies along the semantic continuum from planning to realization. Built upon this foundation, Evo consists of a time-conditioned latent generator and a shared vector field $\mathcal{F}_\theta$ that governs semantic evolution. The model is trained via a variational objective that infers both latent embeddings and their associated time-steps, optimizing a tight evidence lower bound (ELBO) over the sequence likelihood. The resulting training regime unifies next-token prediction and score-based denoising within a single differentiable loss, enabling end-to-end learning with theoretical guarantees. As a special case, we show that Evo recovers both standard AR modeling and diffusion objectives, and converges toward maximum likelihood estimation under mild regularity assumptions. 

Empirically, Evo 8B achieves state-of-the-art or highly competitive results on 15 diverse benchmarks, including reasoning (GSM8K, ARC-C), code generation (HumanEval, MBPP), and general language understanding, while maintaining fast inference speed.

\section{Approach}

\subsection{AR-Diffusion duality: Two Sides of the Same Coin from an evolving perspective}
\label{sec:method_ar_diffusion}

We propose a unified framework for explicit generative modeling, in which both autoregressive models and diffusion models are interpreted as structured paths over latent states governed by probabilistic transitions. 

Formally, let $\mathbf{x} \in \mathbb{R}^d$ be a data sample drawn from the unknown distribution $p_{\text{data}}(\mathbf{x})$. We define a latent Markov path $\mathbf{z}^{(0)}, \mathbf{z}^{(1)}, \dots, \mathbf{z}^{(L)}$ such that $\mathbf{z}^{(L)} = \mathbf{x}$, and posit that the generative model decomposes as
\begin{equation}
    p_\theta(\mathbf{z}^{(0{:}L)}) = p_0(\mathbf{z}^{(0)}) \prod_{\ell=1}^L p_\theta(\mathbf{z}^{(\ell)} \mid \mathbf{z}^{(\ell-1)}),
    \label{eq:latent_path}
\end{equation}
where $p_0$ is a known initialization distribution (e.g., standard Gaussian), and $p_\theta$ parameterizes each transition. The interpretation of the intermediate states $\{\mathbf{z}^{(\ell)}\}$ and the design of transitions determine whether the resulting model behaves like an autoregressive generator or a diffusion sampler.

In the case of autoregressive models, the intermediate states correspond to partial completions of the sequence. For a sequence $\mathbf{x} = (x_1, \dots, x_T)$, we set $\mathbf{z}^{(t)} = (x_1, \dots, x_t)$ and recover the familiar left-to-right factorization
\begin{equation}
    p_\theta(\mathbf{x}) = \prod_{t=1}^T p_\theta(x_t \mid x_{<t}),
\end{equation}
which corresponds to a deterministic, monotonic path in the space of prefix sequences. This process can be lifted into a continuous-time representation by interpolating the discrete sequence trajectory into a smooth path governed by an ODE. Specifically, we define a function $\mathbf{x}_s$ over $s \in [0,1]$, where $\mathbf{x}_0 = \mathbf{0}$ and $\mathbf{x}_1 = \mathbf{x}$, and assume it evolves according to
\begin{equation}
    \frac{d\mathbf{x}_s}{ds} = \mathcal{F}_\theta(\mathbf{x}_s, s),
    \label{eq:ar_flow}
\end{equation}
with $\mathcal{F}_\theta$ being a parameterized vector field that controls how information is revealed along the generation path. This framing renders autoregression as a flow through latent space, aligning with recent formulations of continuous normalizing flows and neural ODEs.

In the case of diffusion models, recall that the forward transition is defined as
\begin{equation}
    q(\mathbf{z}^{(t)} \mid \mathbf{z}^{(t-1)}) = \mathcal{N}(\sqrt{\alpha_t} \mathbf{z}^{(t-1)}, (1 - \alpha_t)\mathbf{I}),
\end{equation}
with a fixed variance schedule $\beta_t = 1 - \alpha_t$. This forward process admits an analytic marginal $q(\mathbf{z}^{(t)} \mid \mathbf{x})$, enabling training of a reverse model $p_\theta(\mathbf{z}^{(t-1)} \mid \mathbf{z}^{(t)})$ that approximates the denoising transition. When discretized with sufficiently small noise steps, the forward process converges to an SDE
\begin{equation}
    d\mathbf{z}_t = \mu(t)\, \mathbf{z}_t\, dt + \sigma(t)\, d\mathbf{w}_t,
\end{equation}
and the corresponding reverse-time dynamics satisfy
\begin{equation}
    d\mathbf{z}_t = \left[-\sigma(t)^2 \nabla_{\mathbf{z}} \log p_t(\mathbf{z})\right] dt + \sigma(t) d\bar{\mathbf{w}}_t,
\end{equation}
where $p_t$ is the time-marginal density and $d\bar{\mathbf{w}}_t$ is a reverse-time Brownian motion. Removing the stochastic term yields the deterministic probability flow ODE
\begin{equation}
    \frac{d\mathbf{z}_t}{dt} = -\sigma(t)^2 \nabla_{\mathbf{z}} \log p_t(\mathbf{z}),
    \label{eq:score_flow}
\end{equation}
which is structurally identical to the AR flow in Eq.~\eqref{eq:ar_flow} under a reparameterization of time.

To make this equivalence precise, we define a smooth transformation $s(t) = \int_0^t \sigma(\tau)^2 d\tau$, mapping diffusion time $t \in [0,1]$ to autoregressive scale $s \in [0,1]$. Under this change of variables, both flows trace the same curves in $\mathbb{R}^d$ when the vector field $\mathcal{F}_\theta(\mathbf{x}, s)$ is chosen as the score function $-\nabla_{\mathbf{x}} \log p_s(\mathbf{x})$. This leads to a formal correspondence between autoregressive generation and score-based sampling, differing only in parametrization and directionality.

Considering the optimization procedures of both models. Autoregressive models directly optimize the negative log-likelihood (NLL), while diffusion models minimize a variational lower bound (ELBO) derived from the KL divergence between the forward and reverse paths. The gap between the ELBO and true NLL is given by
\begin{equation}\label{eq:elbo}
\begin{aligned}
\log p_\theta(\mathbf{x})
 - \mathcal{L}_{\text{ELBO}}     % <-- first (short) line
&= \sum_{t=1}^{T} \mathbb{E}_q \Bigl[
    \mathrm{KL} \\
    &\bigl( 
      q(\mathbf{z}^{(t-1)} \mid \mathbf{z}^{(t)}, \mathbf{x})
\\
&
      p_\theta(\mathbf{z}^{(t-1)} \mid \mathbf{z}^{(t)})
    \bigr)
\Bigr].
\end{aligned}
\end{equation}

which vanishes as $\beta_t = \mathcal{O}(1/T)$ and $T \to \infty$, implying that $\mathcal{L}_{\text{ELBO}} \to \mathcal{L}_{\text{AR}}$.

According to the above analysis, we can see that autoregressive and diffusion models share the same underlying principle: both generate data by following a score-guided path through latent space. They differ mainly in how this path is parameterized—AR takes a deterministic, step-by-step route from data to data, while diffusion follows a stochastic reverse path from noise to data. Despite these differences, both can be viewed as Markovian approximations of the same continuous generative process. Since AR and diffusion can be seen as two ends of a continuous spectrum of path-based generative models, where AR excels at deterministic fine-grained refinement and diffusion at stochastic semantic planning, this unified perspective naturally motivates our hybrid design, assigning each model to the stage of generation where it performs best.

\subsection{Progressive Latent Trajectory Modeling}
We propose \textbf{Evo}, a duality latent flow model that generates text by evolving a sequence of latent vectors along a continuous semantic trajectory. Let $X = (x_1, \dots, x_N) \in \mathcal{V}^N$ denote the target token sequence, and let $Z = \{z_i\}_{i=1}^N$, with each $z_i \in \mathbb{R}^d$ representing a latent code for token $x_i$. Generation is modeled as a learned vector field over the latent space, conditioned on a progression variable $t_i \in [0, 1]$ at each position. The variable $t_i$ encodes the semantic maturity of token $x_i$, interpolating between coarse planning ($t_i \approx 1$) and fine-grained realization ($t_i \approx 0$).

We define the generation likelihood via a continuous-time latent flow:
\begin{equation}
p_\Theta(X) = \int p(X \mid Z)\, p_\theta(Z \mid \{t_i\})\, dZ,
\label{eq:uniflow_likelihood}
\end{equation}
where $p_\theta(Z \mid \{t_i\})$ is parameterized by a time-conditioned Transformer that evolves each latent vector $z_i$ along its own trajectory:
\begin{equation}
\frac{d z_i(t)}{dt} = \mathcal{F}_\theta(z_i(t), t_i),
\label{eq:latent_flow}
\end{equation}
with $\mathcal{F}_\theta$ denoting the learned flow field shared across tokens and time. Each $z_i(0)$ is initialized from a prompt or prior distribution, and $z_i(1)$ is mapped to output token $x_i$ via a decoding projection.

The per-token time-step $t_i$ is predicted jointly with the latent vectors and controls both the generation behavior and its computational depth. In practice, we discretize $t_i$ into $K$ refinement steps, with smaller $t_i$ requiring fewer updates and larger $t_i$ enabling more extensive refinement. This enables adaptive, token-wise allocation of generation effort and supports efficient decoding with fidelity-aware early stopping.

By continuously parameterizing both the generation dynamics and the delegation from planning to refinement via $t_i$, Evo eliminates the need for discrete scaffolds or hard stage transitions. The result is a fully differentiable, flow-based generation process that smoothly integrates semantic intent and linguistic precision within a single unified model.

\subsection{Training Objective}

Evo is trained to maximize the log-likelihood of sequences $X \sim p^*$ under a continuous latent trajectory model. Since exact marginalization over latent paths $Z = \{z_i\}_{i=1}^N$ and progression variables $\{t_i\}_{i=1}^N$ is intractable, we optimize a variational lower bound:
\begin{equation}\tag{2}
\begin{aligned}
\log p_{\Theta}(X)
&\ge \mathbb{E}_{Z,\, t \sim q(Z,t\mid X)}\Bigl[
    \log p_\theta(Z,t) 
\\
&\qquad\qquad + \log p(X\mid Z)\\
   &\qquad\qquad - \log q(Z,t\mid X)
\Bigr].
\end{aligned}
\end{equation}
where $q(Z, t \mid X)$ is an approximate posterior that infers both latent vectors and their associated generation times. The model $p_\theta(Z, t)$ defines a time-conditioned generative flow, while $p(X \mid Z)$ decodes the final latent states into tokens.

At each position $i$, the model predicts a latent vector $z_i$ evolving along a learned trajectory with time-step $t_i \in [0, 1]$. We discretize the trajectory into $K$ steps and train the model to predict the next latent state under the learned vector field $\mathcal{F}_\theta$:
\begin{equation}\tag{3}
\begin{aligned}
\mathcal{L}_{\text{Flow}}(\theta) 
&= \sum_{t=0}^{K-1} \mathbb{E}_{\substack{X \sim p^*\\ Z,\, t \sim q(Z,t\mid X)}}\\
&\Bigl[
    - \log p_\theta\!\left(z^{(t+1)} \mid z^{(t)}, t\right)
\Bigr].
\end{aligned}
\end{equation}

where $z^{(0)}$ is initialized from the prompt or prior, and the flow steps are truncated for each token based on $t_i$. This loss generalizes both next-token prediction (at $t_i \approx 0$) and diffusion denoising (at $t_i \approx 1$), with all intermediate behaviors supported via the continuous time signal. We parameterize $t_i$ as a continuous, learnable variable that reflects the confidence or semantic maturity of the latent vector $z_i$. The posterior $q(Z, t \mid X)$ jointly predicts $z_i$ and $t_i$ from the ground-truth sequence. During training, we estimate $t_i$ via backpropagation to optimize the full ELBO.

We optionally initialize Evo from pretrained language models by mapping their token embeddings to latent initializations. Training can proceed end-to-end over both latent flows and time-step predictors, or with frozen early-stage priors. The model supports parameter sharing across all positions and times, enabling unified modeling of semantic planning and linguistic realization in a single trajectory-based framework.

\subsection{Implementation Details}

\paragraph{Architecture.}  
Evo is implemented as a single Transformer-based sequence model with shared parameters across the entire generation trajectory. Each token position is associated with a continuous latent vector $z_i \in \mathbb{R}^d$ and a learned progression time-step $t_i \in [0, 1]$. The model operates in a time-conditioned fashion: at each refinement step, it receives $\{z_i^{(t)}, t_i\}$ as input and updates the latent trajectory using a learned vector field $\mathcal{F}_\theta$. The Transformer incorporates $t_i$ into the computation via sinusoidal time embeddings added to each token’s hidden state, allowing the model to modulate behavior across the coarse-to-fine spectrum. Prompts are encoded as standard embeddings prepended to the sequence. We adopt a decoder-only architecture but augmented to support full-sequence latent flow updates instead of purely causal decoding. Positional encodings and cross-token dependencies are preserved across time steps via full self-attention. The model supports flexible refinement schedules and can interpolate seamlessly between autoregressive and denoising behavior via the learned $t_i$ values.

\paragraph{Refinement Schedule.}  
Generation proceeds over a fixed number of refinement steps $K_{\max}$ (typically 20), where each latent vector $z_i$ is updated according to its learned time-step $t_i$. Rather than masking tokens or injecting discrete corruption, we use continuous noise perturbation during training. Specifically, we add Gaussian noise to the latent embeddings proportional to a scheduled diffusion time $t$, and train the model to denoise and reconstruct the true target embeddings. At each step $t$, the update rule follows a discretized ODE formulation:
\[
z_i^{(t+1)} = z_i^{(t)} + \Delta t \cdot \mathcal{F}_\theta(z_i^{(t)}, t_i),
\]
where $\Delta t$ is the step size and $\mathcal{F}_\theta$ is the shared time-conditioned flow field. For tokens with lower $t_i$, updates converge quickly; for higher $t_i$, refinement proceeds across more steps. In practice, we optionally truncate updates per token after $K_i = \lfloor K_{\max} \cdot t_i \rfloor$ to save computation, though full trajectories remain supported for all tokens.

\paragraph{Sampling.}  
During inference, the model first samples initial latent states $z_i^{(0)}$ and progression steps $t_i$ for each token using top-$p$ nucleus sampling ($p = 0.9$, temperature $0.7$) over the latent prior distribution. These samples define the initial semantic sketch and generation depth per token. The model then refines the latent trajectory deterministically over $K_{\max}$ steps without adding noise, producing stable and coherent generations. The final token sequence is decoded via a linear projection or nearest neighbor search over a pretrained embedding space.

\section{Experiments}

\subsection{Experimental Setting}
\label{sec:exp_setting}

\paragraph{Training Setup.}
We train \textbf{Evo} at five parameter scales from 1.3B to 13B, and report all main results using the XL configuration (8.5B). Model depth and width are scaled proportionally across sizes. Evo is a decoder-only Transformer augmented with a continuous progression variable \(t_i \in [0,1]\) and a shared latent flow with full-sequence self-attention. Pretraining is conducted on a 3.4T-token multi-domain corpus with a maximum context length of 4K tokens, using a 64K SentencePiece BPE tokenizer. Optimization follows standard large-scale LLM training with AdamW, cosine learning rate scheduling, bfloat16 mixed precision, and DeepSpeed ZeRO-3 on NVIDIA A100 GPUs. After pretraining, Evo is supervised fine-tuned using the same post-training dataset mixture and training settings as LLaDA \cite{zhu2025llada}. Further details are provided in the Appendix.

\paragraph{Baselines, and Evaluation.}
We evaluate Evo 8B against recent open-source models of comparable scale, including AR-only (LLaMA3 8B \cite{dubey2024llama}, Qwen2.5 7B \cite{hui2024qwen2}), diffusion-only (LLaDA 8B \cite{zhu2025llada}, MDLM 7B \cite{sahoo2024simple}), and AR+diffusion hybrids (BD3-LM 7B \cite{arriola2025block}, ARD 7B \cite{wu2023ar}). The benchmark suite covers general understanding and reasoning, mathematics and science, code generation, and Chinese language understanding. For LLaMA3 and Qwen2.5, we initialize from their publicly released pretrained checkpoints and apply the same post-training procedure. Other diffusion-based and hybrid baselines are trained following their respective papers and code, with training data sources, tokenization, and evaluation protocols matched to our setup. All models are evaluated under identical prompts, in-context learning settings, and decoding configurations. Code tasks use deterministic decoding (temperature 0) and pass@1 accuracy, while reasoning and knowledge tasks report final-answer accuracy. Inference speed is measured in tokens per second on a single NVIDIA A100 GPU, including prompt encoding time. Further details are provided in the Appendix.

\newcommand{\capbox}[2]{%
  \begingroup
  \setlength{\fboxsep}{1pt}% padding
  \colorbox{#1}{\strut #2}%
  \endgroup
}

\begin{table*}[t]
\centering
\small
\caption{
Benchmark Results of Pre-trained LLMs.
\capbox{ardiff}{Evo 8B},
\capbox{ardiff}{BD3-LM 7B},
\capbox{ardiff}{ARD 7B} (AR + Diffusion hybrid),
\capbox{diff}{LLaDA 8B},
\capbox{diff}{MDLM 7B} (Diffusion-based),
\capbox{aronly}{LLaMA3 8B},
and \capbox{aronly}{Qwen2.5 7B} (AR-based)
are evaluated under the same protocol.
The numbers in parentheses represent the number of shots.
}
\vspace{-5pt}

\resizebox{\linewidth}{!}{
\begin{tabular}{l|ccc|cc|cc}

\toprule
Model
& \cellcolor{ardiff}Evo 8B
& \cellcolor{ardiff}BD3-LM 7B
& \cellcolor{ardiff}ARD 7B
& \cellcolor{diff}LLaDA 8B
& \cellcolor{diff}MDLM 7B
& \cellcolor{aronly}LLaMA3 8B
& \cellcolor{aronly}Qwen2.5 7B \\ \midrule

E2E Latency (s)
& 8.6
& 14.2
& 32.5
& 21.8
& 18.9
& 7.4
& 8.1 \\

Inference Speed (tokens/s)
& 52
& 28
& 12
& 16
& 22
& 58
& 46 \\

\midrule
\multicolumn{8}{l}{\textbf{General Tasks}} \\

MMLU & \textbf{76.8} (5) & 70.0 (5) & 46.2 (5) & 65.6 (5) & 64.0 (5) & 65.1 (5) & 74.5 (5) \\
BBH & 68.4 (3) & 61.8 (3) & 36.9 (3) & 49.6 (3) & 55.8 (3) & 57.9 (3) & \textbf{70.1} (3) \\
ARC-C & \textbf{65.6} (25) & 60.2 (25) & 46.7 (0) & 48.2 (0) & 59.6 (25) & 52.8 (0) & 64.0 (25) \\
HellaSwag & 82.1 (0) & \textbf{83.0} (10) & 75.6 (0) & 72.2 (0) & 81.2 (10) & 78.8 (0) & 80.5 (10) \\
TruthfulQA & \textbf{58.1} (0) & 53.8 (0) & 39.4 (0) & 46.8 (0) & 42.0 (0) & 44.5 (0) & 56.7 (0) \\
WinoGrande & 76.3 (5) & \textbf{78.1} (5) & 72.2 (5) & 75.1 (5) & 77.4 (5) & 77.6 (5) & 75.6 (5) \\
PIQA & \textbf{81.2} (0) & - & 79.4 (0) & 74.7 (0) & - & 80.3 (0) & - \\

\midrule
\multicolumn{8}{l}{\textbf{Mathematics \& Science}} \\

GSM8K & \textbf{86.4} (4) & 79.8 (4) & 15.1 (4) & 70.9 (4) & 36.5 (4) & 52.7 (4) & 85.8 (4) \\
Math & \textbf{54.9} (4) & 42.9 (4) & 3.6 (4) & 27.0 (4) & 10.0 (4) & 15.4 (4) & 50.2 (4) \\
GPQA & \textbf{38.4} (5) & 30.5 (5) & 25.3 (5) & 26.4 (5) & 24.9 (5) & 26.2 (5) & 36.1 (5) \\

\midrule
\multicolumn{8}{l}{\textbf{Code}} \\

HumanEval & \textbf{60.6} (0) & 50.6 (0) & 13.4 (0) & 33.2 (0) & 29.0 (0) & 34.6 (0) & 58.3 (0) \\
HumanEval-FIM & 73.4 (2) & 71.8 (0) & 27.2 (2) & \textbf{74.0} (2) & - & 73.0 (2) & - \\
MBPP & \textbf{77.4} (4) & 63.7 (0) & 18.9 (4) & 38.5 (4) & 51.3 (0) & 47.1 (4) & 75.2 (0) \\

\midrule
\multicolumn{8}{l}{\textbf{Chinese}} \\

CMMLU & \textbf{88.1} (5) & 84.2 (5) & 33.1 (5) & 70.2 (5) & 47.0 (5) & 50.4 (5) & - \\
C-Eval & 80.4 (5) & \textbf{82.7} (5) & 34.6 (5) & 70.1 (5) & 45.4 (5) & 52.0 (5) & - \\ \bottomrule
\end{tabular}}
\label{tab:llm-benchmark}
\vspace{-10pt}
\end{table*}

\subsection{Main Results}
We present our results in Tab.~\ref{tab:llm-benchmark} (pre-trained) and Tab.~\ref{tab:llm-posttrain} (post-trained), together with end-to-end latency and decoding throughput measured under a unified single-request inference setting.

\begin{table*}[t]
\centering
\small
\caption{Benchmark Results of Post-trained LLMs. \textbf{Evo 8B} and LLaDA 8B post-trained via SFT. The numbers in parentheses represent the number of shots.}
\vspace{-5pt}
\resizebox{\linewidth}{!}{
\begin{tabular}{l|ccc|cc|cc}

\toprule
Model
& \cellcolor{ardiff}Evo 8B
& \cellcolor{ardiff}BD3-LM 7B
& \cellcolor{ardiff}ARD 7B
& \cellcolor{diff}LLaDA 8B
& \cellcolor{diff}MDLM 9B
& \cellcolor{aronly}LLaMA3 8B
& \cellcolor{aronly}Qwen2.5 7B \\ \midrule

E2E Latency (s)
& 8.6
& 14.2
& 32.5
& 21.8
& 18.9
& 7.4
& 8.1 \\

Inference Speed (tokens/s)
& 52
& 28
& 12
& 16
& 22
& 58
& 46 \\

Post-training & SFT & SFT+RL & SFT+RL & SFT & SFT+RL & SFT+RL & SFT+RL \\

\midrule
\multicolumn{8}{l}{\textbf{General Tasks}} \\

MMLU
& \textbf{78.6} (5)
& 75.1 (5)
& 48.9 (5)
& 65.5 (5)
& 70.2 (5)
& 68.4 (5)
& 73.8 (5) \\

MMLU-pro
& \textbf{57.2} (0)
& 48.1 (5)
& 12.8 (0)
& 37.0 (0)
& 52.1 (5)
& 41.9 (0)
& 56.3 (5) \\

Hellaswag
& \textbf{86.4} (0)
& 84.8 (0)
& 50.2 (0)
& 74.6 (0)
& -
& 75.5 (0)
& 82.2 (0) \\

ARC-C
& \textbf{92.5} (0)
& 86.7 (0)
& 56.9 (0)
& 88.5 (0)
& -
& 82.4 (0)
& - \\

\midrule
\multicolumn{8}{l}{\textbf{Mathematics \& Science}} \\

GSM8K
& 89.3 (4)
& 85.7 (0)
& 31.0 (4)
& 78.6 (4)
& 76.7 (0)
& 78.3 (4)
& \textbf{91.6} (0) \\

Math
& \textbf{78.8} (0)
& 74.3 (0)
& 15.6 (0)
& 26.6 (0)
& 44.3 (0)
& 29.6 (0)
& 75.5 (0) \\

GPQA
& \textbf{39.1} (5)
& 34.3 (0)
& 24.2 (5)
& 31.8 (5)
& 32.8 (0)
& 31.9 (5)
& 36.4 (0) \\

\midrule
\multicolumn{8}{l}{\textbf{Code}} \\

HumanEval
& \textbf{86.7} (0)
& 82.9 (0)
& 16.5 (0)
& 47.6 (0)
& 68.9 (0)
& 59.8 (0)
& 84.8 (0) \\

MBPP
& \textbf{81.1} (4)
& 67.2 (0)
& 20.6 (4)
& 34.2 (4)
& 74.9 (0)
& 57.6 (4)
& 79.2 (0) \\
\bottomrule
\end{tabular}}
\label{tab:llm-posttrain}
\vspace{-10pt}
\end{table*}

\paragraph{Performance.}
Evo 8B achieves the strongest and most balanced performance across all capability categories, with its largest margins appearing on tasks that require explicit global planning followed by precise constraint satisfaction. In the pre-trained setting, Evo leads on GSM8K (86.4), MATH (54.9), GPQA (38.4), HumanEval (60.6), and MBPP (77.4), substantially outperforming AR-only baselines such as LLaMA3 on MATH (+24.0) and HumanEval (+25.0). These gains indicate that strictly sequential AR decoding struggles once early reasoning errors are committed. Compared to the strongest AR+diffusion hybrid baseline BD3-LM, Evo further improves GSM8K (+6.6), MATH (+12.0), GPQA (+7.9), and HumanEval (+10.0), suggesting that fixing the AR–diffusion boundary at the block level limits the model’s ability to refine high-level reasoning decisions. After post-training, the same pattern persists: Evo reaches 92.5 on ARC-C, 78.8 on MATH, and 86.7 on HumanEval, yielding a more uniform capability profile than models that peak on isolated benchmarks (e.g., Qwen2.5 on GSM8K).

\paragraph{Efficiency.}
These performance gains are achieved without incurring the high inference cost typically associated with diffusion-based generation because Evo restricts diffusion-style computation to semantically uncertain regions rather than applying it uniformly across tokens or blocks. As a result, Evo operates at near--AR decoding speed (52 tokens/s, 8.6\,s end-to-end latency), closely matching LLaMA3 (58 tokens/s, 7.4\,s) and Qwen2.5 (46 tokens/s, 8.1\,s), while substantially outperforming other AR--diffusion hybrids and diffusion-only models in latency. In contrast, BD3-LM performs mandatory diffusion refinement at the block level, introducing additional model evaluations even when large portions of the sequence are already semantically resolved; this rigidity manifests as higher latency (14.2\,s). AR-Diffusion further amplifies this inefficiency by enforcing position-dependent denoising schedules that allocate the most refinement steps to later tokens regardless of semantic certainty, leading to both high latency (32.5\,s). Diffusion-only models such as LLaDA and MDLM incur even larger latency penalties (21.8\,s and 18.9\,s, respectively) because iterative denoising is applied across the entire sequence.

Together, these results show that Evo’s end-to-end duality, which learns to allocate Diffusion-style global planning and AR-style local refinement along a shared decoding trajectory—enables early commitment to low-entropy structure where necessary and selective refinement only where uncertainty remains. This adaptive coordination directly translates into stronger reasoning and code performance and practical inference efficiency, improving the accuracy–efficiency frontier over both rigid hybrids and diffusion-dominant approaches.

\subsection{Scalability}
\begin{figure}[h]
\centering
\includegraphics[width=\linewidth]{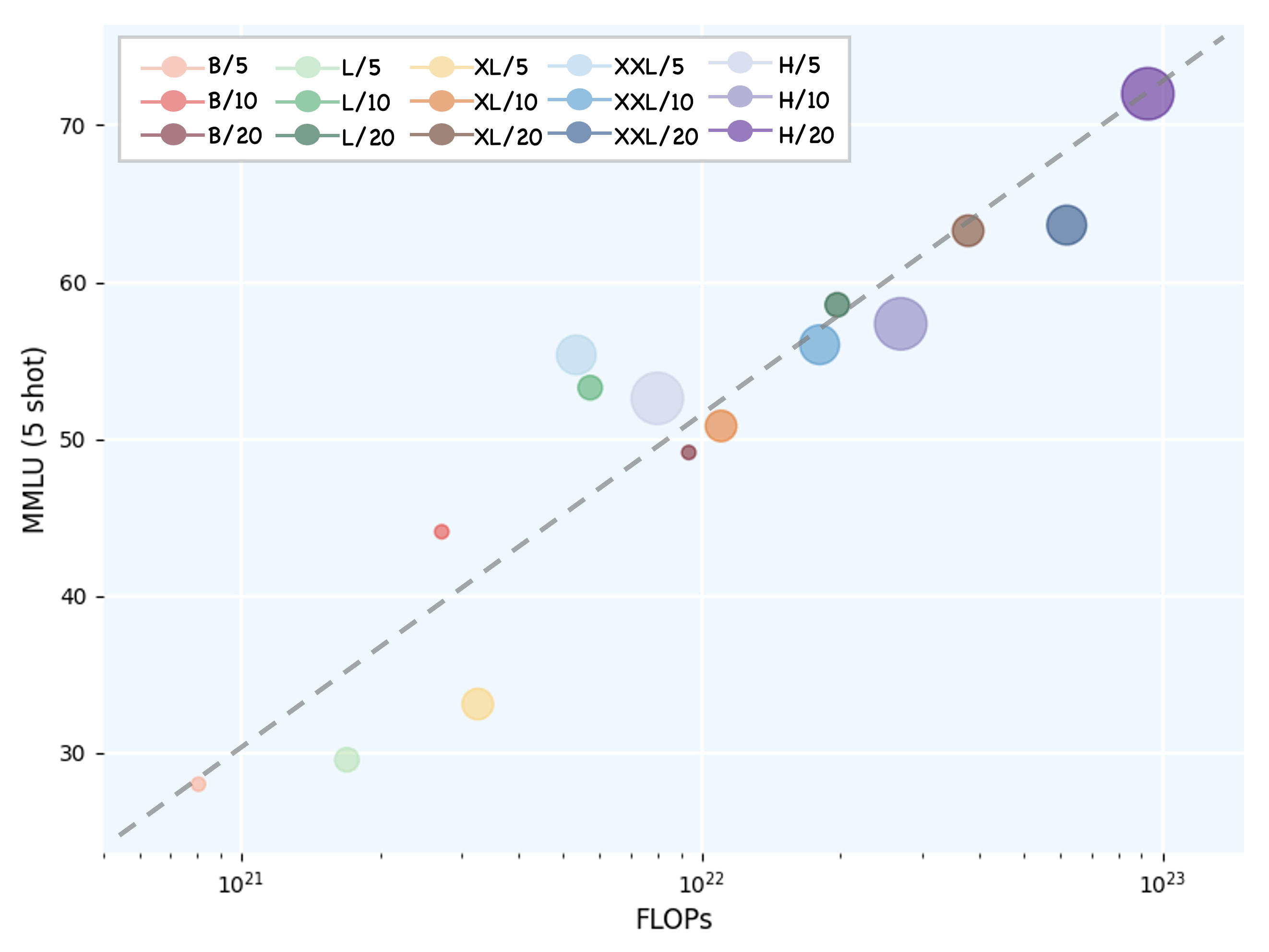}
\vspace{-20pt}
\caption{The scaling capability of Evo on MMLU.}
\label{fig:scale}
\vspace{-15pt}
\end{figure}

We investigate the scaling properties of Evo by varying model size and refinement steps. We train 15 variants across five model sizes (S, B, L, XL, H) and three refinement schedules ($K \in \{5, 10, 20\}$), denoted as Evo-\{size\}/\{steps\}. Model sizes range from S (1.3B parameters, 24 layers, 1024d latents, 16 heads) to H (13B parameters, 48 layers, 2560d latents, 40 heads), with intermediate sizes B (2.7B), L (5.2B), and XL (8.5B) following proportional scaling of depth and width. Each base architecture is evaluated with 5, 10, and 20 refinement steps to study the interplay between model capacity and iterative refinement depth. This design spans computational budgets from around $10^{21}$ to $10^{23}$ FLOPs. In Fig. \ref{fig:scale}, we observe robust correlation observed between model Flops and performance. This implies that Evo has good scaling capability given more computations.

\subsection{Ablation Study}
We conduct ablations to validate Evo’s design, summarized in Fig~\ref{fig:ablation} and Fig~\ref{fig:ablation_arch}. Evaluations cover perplexity (PPL), average task accuracy (MMLU, GSM8K), and refinement steps per token.

\paragraph{Effect of Time Conditioning.}
Removing time conditioning severely hurts PPL and accuracy, showing its role in modulating generation along a continuous trajectory. Fixed time-steps also underperform: $t_i=0$ (pure AR) yields the worst accuracy, $t_i=1$ (pure diffusion) improves fluency but is inefficient, and $t_i=0.5$ gives a trade-off but remains inferior to learned time.

\paragraph{Learned vs. Heuristic Steps.}
A heuristic based on AR entropy performs reasonably but lags behind learned $t_i$, indicating that confidence signals alone are insufficient and direct end-to-end optimization is essential.

\paragraph{Architecture and Pretraining.}
Replacing Evo’s unified latent flow with separate AR and diffusion modules lowers accuracy and raises PPL, highlighting the value of joint modeling. Removing shared embeddings weakens alignment across steps, and dropping pretrained initialization slows convergence and reduces accuracy. By contrast, freezing the AR backbone retains most performance, showing Evo’s compatibility with modular adaptation.

Overall, Evo’s gains arise from its time-conditioned latent flow, unified architecture, and shared representations.

\begin{figure}[t]
\centering
\small
\begin{tabular}{lcc}
\toprule
\textbf{Ablation Variant} & \textbf{PPL ↓} & \textbf{Task Acc ↑} \\
\midrule
\textit{w/o Time Conditioning} & 9.6 & 69.1 \\
\textit{Fixed $t_i = 0$ (AR-only)} & 10.2 & 69.6 \\
\textit{Fixed $t_i = 1$ (Diff-only)} & 8.6 & 77.0 \\
\textit{Fixed $t_i = 0.5$ (Uniform)} & 8.8 & 77.9 \\
\textit{Heuristic $t_i$ (entropy)} & 8.4 & 78.2 \\
\midrule
\textbf{Evo (full)} & \textbf{8.1} & \textbf{81.6} \\
\bottomrule
\end{tabular}
\vspace{-5pt}
\caption{Ablation study of Evo on text modeling and reasoning tasks. PPL: perplexity on validation set; Task Acc: average task accuracy (MMLU and GSM8K). Results highlight the importance of time conditioning, learned token-wise $t_i$, and adaptive refinement.}
\label{fig:ablation}
\end{figure}

\begin{figure}[t]
\centering
\small
\begin{tabular}{lcc}
\toprule
\textbf{Ablation Variant} & \textbf{PPL ↓} & \textbf{Task Acc ↑} \\
\midrule
\textit{Two-Module (AR + Diff)} & 8.6 & 77.9 \\
\textit{No Shared Embeddings} & 8.4 & 78.3 \\
\textit{No Pretrained Init} & 9.1 & 76.2 \\
\textit{Frozen AR Backbone} & 8.7 & 80.4 \\
\midrule
\textbf{Evo (full)} & \textbf{8.1} & \textbf{81.6} \\
\bottomrule
\end{tabular}
\vspace{-5pt}
\caption{Ablation on architecture and pretraining. Evo benefits from a unified model, embedding sharing, and pretrained initialization. Task accuracy is averaged across MMLU and GSM8K.}
\label{fig:ablation_arch}
\end{figure}

\section{Related Work}
\subsection{Autoregressive Language Models and Diffusion-Based Language Models}
Autoregressive models generate text by factorizing the joint probability of a sequence into a product of conditional probabilities, typically producing tokens sequentially from left to right. Transformer-based architectures, such as GPT \citep{radford2018gpt}, GPT-2 \citep{radford2019gpt2}, and subsequent large-scale models like GPT-3 \citep{brown2020gpt3}, have demonstrated remarkable success in generating coherent and contextually relevant text. These models rely on causal masking to enforce sequential dependencies, enabling efficient training and sampling. However, AR models often struggle with global coherence over long sequences and lack mechanisms for iterative refinement

Diffusion models, originally popularized in image generation \citep{ho2020denoising, song2020score}, have recently been adapted for text generation. These models operate by iteratively denoising a noisy sequence, typically starting from a random noise distribution and refining it toward a target sequence. Works such as D3PM \citep{austin2021d3pm} and Diffusion-LM \citep{li2022diffusion} have explored discrete diffusion processes for text, modeling token distributions over a series of denoising steps. Continuous diffusion approaches, such as those proposed by \citet{song2020score}, have also been extended to text by embedding tokens in a continuous space \citep{gong2022diffuseq}.

\subsection{Autoregressive and Diffusion Hybrid Models}
Recent work has explored combining the complementary strengths of autoregressive (AR) and diffusion-based generation for text, aiming to balance causal decoding with iterative refinement. Early hybrid attempts include step-unrolled denoising autoencoders \cite{savinov2021step}, which integrate denoising objectives into a sequential generation process, and flow-based text models \cite{hoogeboom2021argmax, chen2023latentflow}, which leverage continuous normalizing flows to bridge discrete token spaces and continuous latent dynamics. While these approaches demonstrate that diffusion-style refinement can be reconciled with sequential generation, they typically rely on separate training objectives or predefined inference schedules, and do not explicitly model how global semantic structure should stabilize and refined during generation.

More recent work has focused on tighter AR–diffusion hybrids with explicit causal structure. AR-Diffusion (ARD) \cite{wu2023ar} extends SSD-style diffusion \cite{han2022ssd}, \cite{han2023ssd2} by imposing a left-to-right noise schedule, ensuring causal dependency while retaining diffusion-based denoising. Later, BD3-LM \cite{arriola2025block} adopts a discrete block diffusion formulation, generating blocks autoregressively while applying diffusion-based refinement within each block. This strategy improves efficiency over fully masked diffusion, but fixes the interaction between planning and refinement at the block level, constraining the model’s ability to adapt refinement granularity dynamically within a sentence. 

\section{Conclusion}
In this work, we introduced Evo, a unified latent flow model that bridges AR and diffusion paradigms through a continuou trajectory. Evo integrates confident planning and fine-grained refinement within a single architecture. Our theoretical analysis established AR and diffusion as discretizations of the same probability flow, and our empirical results demonstrated that Evo achieves SOTA performance across diverse coding and knowledge benchmarks while maintaining fast decoding.
\clearpage

\section{Limitations}
While Evo demonstrates strong performance and a favorable accuracy--efficiency trade-off, it also has several limitations that warrant discussion.

\paragraph{Training Cost.}
Evo introduces additional training complexity compared to standard autoregressive or diffusion-only models, as it must jointly learn to evolve a semantic scaffold and coordinate autoregressive commitment with diffusion-based refinement along a shared latent trajectory. Although we adopt efficient training strategies to amortize this cost, Evo training remains more expensive than pure AR training and moderately more costly than standard diffusion training. Further optimization of the training objective, curriculum, and scheduling may be required to reduce this overhead, particularly for larger model scales.

\paragraph{Inference Efficiency Sensitivity.}
Although Evo avoids the high inference cost typical of diffusion-based generation, its decoding efficiency depends on how semantic uncertainty is distributed across tokens. Tasks that require prolonged refinement over large portions of the sequence may reduce the efficiency advantage over AR-only models. In addition, while Evo mitigates the rigid scheduling constraints of prior AR--diffusion hybrids, the optimal balance between planning and refinement can still be task-dependent, and suboptimal allocation may affect either latency or output quality.

\paragraph{Scalability and Controllability.}
Evo generates text by evolving a shared semantic scaffold rather than committing strictly left-to-right, which introduces new challenges for controllability and interpretability. While this flexibility enables stronger reasoning, it may complicate fine-grained control over generation attributes such as length, especially in long-form generation. Understanding how to best guide or constrain scaffold evolution remains an open problem.

\paragraph{General Generative Model Risks.}
Like other large generative language models, Evo is subject to well-known limitations, including hallucination, exposure to training data biases, potential copyright infringement, and the risk of generating harmful or misleading content. Addressing these issues requires complementary techniques such as alignment training, filtering, and post-hoc safety mechanisms, which are orthogonal to the core modeling contributions of this work.

Overall, while Evo advances the integration of autoregressive and diffusion-based generation, these limitations highlight important directions for future research in training efficiency, inference control, and safe deployment.

\bibliography{acl_latex}

\clearpage

\section{Appendix}
\subsection{Training Details}

We trained Evo across five model scales, ranging from 1.3B to 13B parameters, with the XL configuration (8.5B parameters) used for the main experiments. Architectural hyperparameters were scaled proportionally in depth and width to maintain consistent capacity-to-compute ratios. Evo extends a standard decoder-only Transformer by introducing a continuous progression time embedding $t_i \in [0, 1]$ that is injected into each token representation via a sinusoidal projection. A shared vector field $F_\theta$ is applied across all tokens and refinement steps, updating latent states in a time-conditioned manner. Latent flow updates replace purely causal masking while preserving inter-token dependencies through full-sequence self-attention. At every refinement step, the model attends over the entire latent sequence, modulating generation behaviour along the coarse-to-fine spectrum dictated by $t_i$.

The pretraining corpus consists of 3.4 trillion tokens drawn from a mixture of domains. English text, accounting for roughly sixty percent of the data, is sourced from filtered Common Crawl, Wikipedia, books, StackExchange, ArXiv, and code repositories. Multilingual data covers more than thirty languages with high-quality Wikipedia and curated web sources, contributing about twenty-five percent of the total. Code data comprises approximately ten percent, obtained from permissively licensed GitHub repositories in languages including Python, JavaScript, C/C++, Java, Go, and Rust. Mathematics and science resources, representing the remaining five percent, are drawn from datasets such as ProofNet, MathQA, and scientific publications. Data cleaning includes deduplication with MinHash, language identification filtering, toxicity removal, and heuristic screening for quality. All documents are normalized to a maximum length of 4K tokens for efficient batching.

Tokenization is performed using a SentencePiece BPE vocabulary of 64K merge operations, with case-sensitive encoding and special tokens for sequence control and model operations, including a dedicated token to explicitly condition on progression time. This facilitates training regimes where $t_i$ can be externally specified or learned.

The training objective is a unified variational lower bound that combines three components: a latent flow prediction loss, a standard autoregressive negative log-likelihood term, and a score-matching loss for denoising. The autoregressive term dominates when $t_i$ is close to zero, while the score-matching term dominates as $t_i$ approaches one. We set the weights of the autoregressive and diffusion losses to unity and anneal their relative importance during warmup, allowing the latent flow loss to govern later training. The optimizer is AdamW with $\beta_1=0.9$, $\beta_2=0.95$, $\epsilon=10^{-8}$, and weight decay of 0.1. A cosine learning rate schedule with three thousand warmup steps is used, with peak learning rates of $2\times 10^{-4}$ for smaller models and $1.5\times 10^{-4}$ for larger ones. Batch sizes correspond to two million tokens for smaller models and four million for larger ones, with a fixed sequence length of 4K tokens and gradient accumulation employed to fit the model within available GPU memory.

Noise is injected into the latent space using a continuous Gaussian perturbation $z_i^{(t)} = \sqrt{\alpha_t} z_i^{(0)} + \sqrt{1-\alpha_t}\,\epsilon$, where $\alpha_t$ follows a cosine schedule $\alpha_t = \cos^2\!\left(\frac{\pi}{2} t\right)$. This allows smooth interpolation between deterministic autoregressive behaviour and stochastic diffusion-like refinement. During inference, noise is disabled and generation proceeds deterministically along the learned ODE path.

Progression times $t_i$ are predicted by a learned head applied to the initial latent vector of each token, consisting of a two-layer MLP with GELU activation and a final sigmoid mapping to $[0,1]$. An entropy regularizer encourages diverse $t_i$ usage across the sequence. The per-token update depth $K_i$ is defined as $\lfloor K_{\max} \cdot t_i \rfloor$, with $K_{\max}$ set to 20 in pretraining. Tokens with smaller $t_i$ values converge quickly with few updates, while larger values permit extended refinement. This mechanism enables adaptive allocation of computation to uncertain tokens.

Regularization techniques include dropout of 0.1 on residual and feed-forward connections, label smoothing of 0.1 for the autoregressive loss, gradient clipping at a global norm of 1.0, and exponential moving averages of model weights with decay $0.9999$ to stabilize inference. Mixed-precision training is performed in bfloat16 for improved efficiency.

Pretraining is conducted on large-scale GPU clusters, with smaller models trained on sixty-four NVIDIA A100 40GB GPUs and larger ones on up to two hundred and fifty-six NVIDIA A100 80GB GPUs. DeepSpeed ZeRO-3 with pipeline parallelism and fully sharded data parallelism is used to minimize memory overhead. Sequence parallelism enables efficient handling of long contexts. Training wall-clock times range from two weeks for the smallest models to seven weeks for the largest.

Post-training consists of supervised fine-tuning on curated instruction-following datasets, reasoning corpora, and code generation benchmarks. The fine-tuning learning rate is set to $5\times 10^{-5}$ with cosine decay and one thousand warmup steps. Batch sizes are set to 512 sequences of 2K tokens, and training runs for 30K–50K steps depending on the domain. For alignment, we experiment with direct preference optimization for helpfulness and harmlessness, as well as self-refinement loops in which the model generates candidate completions, scores them with an internal reward model, and refines outputs via additional deterministic refinement steps.

Evaluation is performed under identical in-context learning settings for all baselines, with both zero-shot and few-shot prompts, as well as chain-of-thought prompting where appropriate. For code benchmarks, we report pass@1 scores using deterministic decoding with temperature zero. Reasoning tasks are evaluated based on accuracy of final answers, and inference speed is measured in tokens per second on a single A100 GPU, including prompt encoding time. This consistent setup ensures fair comparison across autoregressive, diffusion-based, and hybrid architectures, allowing us to isolate the benefits of Evo’s time-conditioned latent flow design.

\subsection{Proof of Proposition}

\noindent\textbf{Proposition 1.} 
\emph{Autoregressive language models and score-based diffusion language models are two discrete parameterizations of the same continuous-time generative process. Specifically, there exists a smooth, monotonic reparameterization of time $s(t)$ such that the deterministic probability flow ODE of a diffusion model coincides with the latent evolution ODE of an autoregressive model, when both are expressed in a continuous latent space.}

\medskip
\noindent\textbf{Proof.}
We begin by recalling the continuous-time latent evolution for an autoregressive (AR) model. Let $x_s$ denote the latent representation of the partially generated sequence at continuous generation time $s \in [0, 1]$, with $x_0$ denoting the start state and $x_1$ the completed sequence. In the continuous limit of left-to-right generation, the dynamics can be written as an ordinary differential equation
\begin{equation}
    \frac{d x_s}{ds} = F_\theta(x_s, s),
    \label{eq:ar_flow2}
\end{equation}
where $F_\theta$ is a deterministic vector field representing the model's generative policy over the latent space.

Now consider a continuous-time score-based diffusion model. Let $z_t$ denote the latent variable at diffusion time $t \in [0,1]$, where $z_1 \sim \mathcal{N}(0, I)$ is pure noise and $z_0$ is the data distribution. The forward noising process can be described by a variance-preserving stochastic differential equation (SDE):
\begin{equation}
    d z_t = \mu(t) z_t \, dt + \sigma(t) \, d w_t,
    \label{eq:forward_sde}
\end{equation}
where $w_t$ is a standard Wiener process. The corresponding reverse-time SDE is given by:
\begin{equation}
    d z_t = \left[ -\mu(t) z_t - \sigma(t)^2 \nabla_z \log p_t(z_t) \right] dt + \sigma(t) \, d \bar{w}_t,
    \label{eq:reverse_sde}
\end{equation}
where $\bar{w}_t$ is reverse-time Brownian motion and $p_t$ denotes the marginal distribution of $z_t$.

Removing the stochastic term from Eq.~\eqref{eq:reverse_sde} yields the deterministic probability flow ODE:
\begin{equation}
    \frac{d z_t}{dt} = -\mu(t) z_t - \sigma(t)^2 \nabla_z \log p_t(z_t).
    \label{eq:prob_flow_ode}
\end{equation}
When the forward process is chosen to be variance-preserving ($\mu(t) = 0$), the first term vanishes and the flow reduces to:
\begin{equation}
    \frac{d z_t}{dt} = -\sigma(t)^2 \nabla_z \log p_t(z_t).
    \label{eq:vp_flow}
\end{equation}

We now establish the equivalence to the AR flow in Eq.~\eqref{eq:ar_flow2}. Define a smooth, strictly increasing reparameterization $s = \phi(t)$ with $\phi(0) = 0$ and $\phi(1) = 1$, given by
\begin{equation}
    s(t) \triangleq \frac{\int_{0}^{t} \sigma(\tau)^2 \, d\tau}{\int_{0}^{1} \sigma(\tau')^2 \, d\tau'}.
    \label{eq:time_reparam}
\end{equation}
Applying the chain rule to Eq.~\eqref{eq:vp_flow} gives
\begin{equation}
    \frac{d z_s}{ds} = \frac{d z_t}{dt} \cdot \frac{dt}{ds} 
    = \left[ -\sigma(t)^2 \nabla_z \log p_t(z_t) \right] \cdot \frac{1}{\sigma(t)^2 / C},
\end{equation}
where $C = \int_{0}^{1} \sigma(\tau')^2 \, d\tau'$ is a positive constant. Simplifying, we obtain
\begin{equation}
    \frac{d z_s}{ds} = - C \, \nabla_z \log p_s(z_s).
    \label{eq:ar_equivalence}
\end{equation}
Eq.~\eqref{eq:ar_equivalence} has the same form as the AR flow in Eq.~\eqref{eq:ar_flow2}, with the identification
\begin{equation}
    F_\theta(x_s, s) \equiv - C \, \nabla_x \log p_s(x_s).
\end{equation}
That is, the deterministic AR evolution is a scaled version of the probability flow ODE for a score-based model, when both are parameterized over the same continuous-time latent trajectory. The AR model can thus be viewed as following the same score field as the diffusion model, but with a different time parametrization and without stochasticity.

Finally, under the standard assumptions that $\sigma(t)$ is continuous and strictly positive, and that $p_t$ admits a differentiable score function $\nabla_z \log p_t(z)$, the mapping in Eq.~\eqref{eq:time_reparam} is smooth and invertible, ensuring the equivalence holds globally over $t \in [0, 1]$. Therefore, autoregressive generation and deterministic diffusion flows are provably two discretizations of a shared continuous-time generative process, completing the proof.

\subsection{Case Studies}

We present qualitative outputs from \textbf{Evo-8B Instruct} to illustrate its hybrid autoregressive–diffusion generation capabilities. The examples demonstrate (1) coherent and fluent text generation in a mixed-step decoding regime, (2) multi-turn dialogue with consistent context retention, and (3) adaptive reasoning that leverages fine-grained token refinement.

\paragraph{Mathematical Reasoning.} 
In Table~\ref{tab:case_math}, Evo correctly solves a multi-step word problem by first drafting an initial reasoning chain and then selectively refining uncertain tokens in later diffusion-style steps.

\begin{table}[h]
\centering
\caption{Mathematical reasoning with adaptive refinement.}
\label{tab:case_math}
\resizebox{\columnwidth}{!}{
\begin{tabular}{p{0.95\columnwidth}}
\toprule
\textbf{User:} Lily can run 12 kilometers per hour for 4 hours. After that, she runs 6 kilometers per hour for another 3 hours. How many kilometers does she run in total? \\
\textbf{Evo:} First, at 12 km/h for 4 hours, Lily covers $48$ kilometers. Then, at 6 km/h for 3 hours, she covers $18$ kilometers. Adding them gives $66$ kilometers in total. \\
\bottomrule
\end{tabular}
}
\end{table}

\paragraph{Multi-Turn Dialogue.} 
Table~\ref{tab:case_dialogue} shows Evo engaging in a multilingual multi-turn conversation. The model maintains dialogue coherence across turns, translating between languages, and composing text under lexical constraints.

\begin{table}[h]
\centering
\caption{Multi-turn dialogue with translation and constrained generation.}
\label{tab:case_dialogue}
\resizebox{\columnwidth}{!}{
\begin{tabular}{p{0.95\columnwidth}}
\toprule
\textbf{User:} Could you share the first two lines of the poem ``The Road Not Taken''? \\
\textbf{Evo:} ``Two roads diverged in a yellow wood, And sorry I could not travel both.'' \\
\textbf{User:} Translate into German. \\
\textbf{Evo:} ``Zwei Wege trennten sich im gelben Wald, und leider konnte ich nicht beide gehen.'' \\
\textbf{User:} Now write a short poem about life choices, each sentence starting with ``C''. \\
\textbf{Evo:} Choices shape the path we walk. Courage guides us through the unknown. Curiosity opens hidden doors. Commitment carries us forward. Clarity shows the way home. \\
\bottomrule
\end{tabular}
}
\end{table}

\paragraph{Step-Adaptive Story Generation.} 
Evo’s hybrid decoding allows for global planning via coarse AR passes followed by localized diffusion refinements. Table~\ref{tab:case_story} shows a short sci-fi story where key sentences are refined for narrative flow.

\begin{table}[h]
\centering
\caption{Step-adaptive story generation. Italics indicate refined sentences.}
\label{tab:case_story}
\resizebox{\columnwidth}{!}{
\begin{tabular}{p{0.95\columnwidth}}
\toprule
\textbf{User:} Write a short science-fiction story about a time-traveling botanist. \\
\textbf{Evo:} Dr. Amina travelled to the year 3050, carrying seeds from the past. She planted them in Martian soil, curious to see what might grow. \textit{Under the twin moons, the seedlings shimmered with an unearthly glow.} The Martian settlers gathered to witness the strange blossoms. \textit{In that moment, Amina realized she had brought not just plants, but hope, across centuries.} \\
\bottomrule
\end{tabular}
}
\end{table}

\paragraph{Multi-Step Reasoning.} 
Evo decomposes complex tasks into intermediate steps, benefiting from its ability to revise earlier parts of the reasoning chain without regenerating the entire sequence.

\begin{table}[h]
\centering
\caption{Mathematical reasoning with token refinement.}
\label{tab:case_math2}
\resizebox{\columnwidth}{!}{
\begin{tabular}{p{0.95\columnwidth}}
\toprule
\textbf{User:} A tank is filled at 15 liters per minute for 8 minutes, then at 10 liters per minute for 12 minutes. How many liters are in the tank? \\
\textbf{Evo:} First stage: $15 \times 8 = 120$ liters. Second stage: $10 \times 12 = 120$ liters. Adding gives $240$ liters in total. \\
\bottomrule
\end{tabular}
}
\end{table}

\paragraph{Multilingual Dialogue.} 
Evo smoothly transitions across languages while retaining context, demonstrating both translation and constrained text generation.

\begin{table}[h]
\centering
\caption{Multi-turn multilingual dialogue.}
\label{tab:case_multi_lang}
\resizebox{\columnwidth}{!}{
\begin{tabular}{p{0.95\columnwidth}}
\toprule
\textbf{User:} What is the proverb ``A stitch in time saves nine'' in German? \\
\textbf{Evo:} German: ``Wer rechtzeitig handelt, erspart sich viel Mühe.'' \\
\textbf{User:} Write a haiku in English using the word ``time''. \\
\textbf{Evo:} Time flows like a stream, \newline Leaves drift in a quiet dream, \newline Dawn breaks, gold and clean. \\
\bottomrule
\end{tabular}
}
\end{table}

\paragraph{Code Completion.} 
Evo’s AR steps quickly produce a syntactically correct draft, while refinement passes fill in missing logic.

\begin{table}[h]
\centering
\caption{Python function generation.}
\label{tab:case_code}
\resizebox{\columnwidth}{!}{
\begin{tabular}{p{0.95\columnwidth}}
\toprule
\textbf{User:} Write a Python function to check if a number is prime. \\
\textbf{Evo:} 
\texttt{def is\_prime(n):} \newline
\quad \texttt{if n < 2: return False} \newline
\quad \texttt{for i in range(2, int(n**0.5)+1):} \newline
\quad\quad \texttt{if n \% i == 0: return False} \newline
\quad \texttt{return True} \\
\bottomrule
\end{tabular}
}
\end{table}

\paragraph{Creative Storytelling.} 
Hybrid decoding enables Evo to generate coherent narratives and then improve phrasing without losing global structure.

\begin{table}[h]
\centering
\caption{Short story with refinement. Italics indicate refined sentences.}
\label{tab:case_creative}
\resizebox{\columnwidth}{!}{
\begin{tabular}{p{0.95\columnwidth}}
\toprule
\textbf{User:} Write a short story about a lost map in the desert. \\
\textbf{Evo:} The wind buried the caravan’s trail in minutes. Karim unfolded the old map, only to find half the ink faded. \textit{The sun blazed overhead, turning sand into a sea of gold.} He followed the faint lines until they ended at a lone acacia tree. \textit{Beneath its roots, a small tin box held the real map, untouched by time.} \\
\bottomrule
\end{tabular}
}
\end{table}

\paragraph{Reversal Reasoning.} 
Unlike pure AR models, Evo treats tokens without strict left-to-right bias, allowing it to recover missing preceding context.

\begin{table}[h]
\centering
\caption{Poem reversal completion.}
\label{tab:case_reversal}
\resizebox{\columnwidth}{!}{
\begin{tabular}{p{0.95\columnwidth}}
\toprule
\textbf{User:} Given the line ``And sorry I could not travel both'', write the preceding line from Robert Frost’s ``The Road Not Taken''. \\
\textbf{Evo:} ``Two roads diverged in a yellow wood.'' \\
\bottomrule
\end{tabular}
}
\end{table}

Across all domains, Evo benefits from its hybrid AR--diffusion decoding in two key ways. First, the AR component ensures rapid generation of a globally coherent structure, making it competitive in speed with pure autoregressive LLMs. Second, the diffusion-style refinement selectively updates uncertain tokens, enabling corrections without regenerating the entire sequence. This is particularly evident in reasoning (Tables~\ref{tab:case_math2}, \ref{tab:case_reversal}) where local numerical or contextual corrections are made after the global plan is laid out. In multilingual dialogue (Table~\ref{tab:case_multi_lang}), the retention of long-range context across turns is aided by the bidirectional attention in refinement steps, which allows earlier utterances to influence later token predictions even in different languages. For creative writing (Table~\ref{tab:case_creative}), refinement passes improve thematic consistency and stylistic flow, showing how Evo can balance efficiency and quality. Finally, in reversal reasoning (Table~\ref{tab:case_reversal}), Evo’s non-directional refinement offers an inherent advantage over strictly causal models, which often struggle when the missing span is at the start of the context.

\end{document}